\documentclass{article}
\usepackage{spconf,amsmath,graphicx}
\usepackage{subcaption}
\usepackage{booktabs}
\usepackage{times}
\usepackage{epsfig}
\usepackage{graphicx}
\usepackage{amsmath}
\usepackage{amssymb}
\usepackage{subcaption}
\usepackage{enumitem}
\usepackage{tablefootnote}
\usepackage{multirow}
\usepackage{capt-of}

\usepackage[table]{xcolor}

\title{A comparison of different atmospheric turbulence simulation methods for image restoration}

\name{Nithin Gopalakrishnan Nair, \hspace{0.1cm} Kangfu Mei\hspace{0.16cm} and \hspace{0.16cm} Vishal M. Patel \thanks{This research is based upon work supported in part by the Ofﬁce of the Director of National Intelligence (ODNI), Intelligence Advanced Research Projects Activity (IARPA), via [2022-21102100005]. The views and conclusions contained herein are those of the authors and should not be interpreted as necessarily representing the ofﬁcial policies, either expressed or implied, of ODNI, IARPA, or the U.S. Government. The US. Government is authorized to reproduce and distribute reprints for governmental purposes notwithstanding any copyright annotation therein.}}
\address{Dept. of Electrical and Computer Engineering, Johns Hopkins University, MD, USA \\\texttt{\{ngopala2, kmei1, vpatel36\}@jhu.edu}}
\begin{document}
 \ninept
\maketitle
\vspace{-0.5cm}
\begin{abstract}
Atmospheric turbulence deteriorates the quality of images captured by long-range imaging systems by introducing  blur and geometric distortions to the captured scene. This leads to a drastic drop in performance when computer vision algorithms like object/face recognition and detection are performed on these images.  In recent years, various deep learning-based atmospheric turbulence mitigation methods have been proposed in the literature.  These methods are often trained using synthetically generated images and tested on real-world images. Hence, the performance of these restoration methods depends on the type of simulation used for training the network.   In this paper, we systematically  evaluate the effectiveness of various turbulence simulation methods on image restoration.  In particular, we evaluate the performance of two state-or-the-art  restoration networks using six simulations method on a real-world LRFID dataset consisting of face images degraded by turbulence.  This paper will provide guidance to the researchers and practitioners working in this field to 
choose the suitable data generation models for training deep 
models for turbulence mitigation. The implementation codes for the simulation methods, source codes for the
networks and the pre-trained models will be publicly made available.

\end{abstract}
\section{Introduction}
\label{sec:intro}


The problem of recognizing a person appearing in an image taken by a long-range imaging system is important in many biometrics and surveillance applications \cite{lau2019restoration,mei2021ltt,yasarla2021learning}.  Atmospheric turbulence adversely degrades the quality of images captured by long-range imaging systems by causing spatially varying blurring effects on each pixel in the captured image. Since most existing face verification algorithms perform best when used on clean images \cite{deng2019arcface,deng2020retinaface}, face verification on such images is performed by first removing the distortions caused due to atmospheric turbulence and then performing verification on the restored images.  Recently some deep learning-based models \cite{lau2019restoration,nair2021confidence,yasarla2021learning,yasarla2020learning,mei2021ltt} have been proposed to tackle this problem of atmospheric turbulence mitigation. These networks are trained using synthetically generated data modelling the turbulence distortion \cite{chak2021subsampled,mei2021ltt}. Although multiple works have previously attempted to model the phenomenon of atmospheric turbulence, there is no detailed analysis of the performance of deep networks trained using these turbulence modeling techniques to restore real-world face images.
 
 Multiple techniques have been proposed in the  literature for modelling atmospheric turbulence \cite{hardie2017simulation, potvin2011simple, mao2021accelerating,chimitt2020simulating,chak2021subsampled,schmidt2010numerical}. Hardie \textit{et al.} \cite{hardie2017simulation}  proposed a numerical wave propagation method for modeling turbulence distortion of a scene under anisoplanatic condition. This method uses an array of Point Spread Functions (PSFs) to perform a spatially varying weighted sum operation on a clean image to generate the turbulence distorted image.
 Potvin \textit{et al.} \cite{potvin2011simple} proposed a simple physical model based on first-order Rytov theory for propagation through turbulence. The method uses two scalar fields derived using a Gaussian and non-isoplanatic PSFs. These scalar fields are further used to model blur and geometric deformations. Bos \textit{et al.} \cite{bos2012technique} utilizes a simplified form of the split-step wave propagation method. Essentially this method employs a series of uniformly spaced phase screens to simulate turbulence distortions. Schmidt \textit{et al.} \cite{schmidt2010numerical} describes a set of wave optics algorithms to simulate atmospheric turbulence. Recently Mao \textit{et al.} \cite{mao2021accelerating}, and Chimitt \textit{et al.} \cite{chimitt2020simulating} have proposed fast turbulence generation algorithms based on computing the tilt and aberration coefficients using Zernike basis functions. Chak \textit{et al.} \cite{chak2021subsampled} decomposes the turbulence mitigation problem into a blurring and geometric distortion problem and model both of these independently. Schwartzman \textit{et al.} \cite{schwartzman2017turbulence} utilizes the physics behind turbulence and generates distortion fields according to the empirical turbulence model. \cite{mei2021ltt} uses elastic deformation operation to model the geometric deformation and Gaussian blur model for the blurring operator. In this paper, we focus on the works by \cite{mao2021accelerating,chimitt2020simulating,mei2021ltt,chak2021subsampled,schwartzman2017turbulence} since they are computationally inexpensive and are suitable for generating data for training deep networks.
   \begin{figure*}[t]
    \centering
    \includegraphics[width=.5\linewidth]{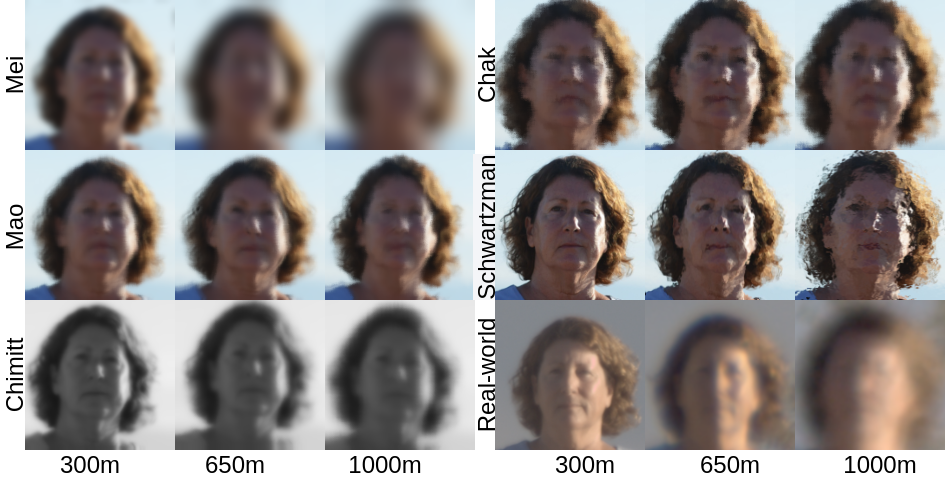}
        \vspace{-0.4cm}
    \caption{Simulated images of ranges 300 m, 650 m and 1000 m.  First row:  Mei \textit{et al.} \cite{mei2021ltt} and Chak \textit{et al.} \cite{chak2021subsampled}. Second row: Mao \textit{et al.} \cite{mao2021accelerating} and Schwartzman \textit{et al.} \cite{schwartzman2017turbulence}. Third row: Chimitt \textit{et al.} \cite{chimitt2020simulating} and real world images. }
    \label{fig:combined}
    \vspace{-0.6cm}
\end{figure*}

The problem of atmospheric turbulence mitigation has also been extensively studied in the literature. Essentially there are two main approaches for mitigating atmospheric turbulence -- adaptive optics-based \cite{roggemann1996imaging, pearson1976atmospheric,tyson2015principles} and image processing-based \cite{anantrasirichai2013atmospheric,anantrasirichai2018atmospheric,aubailly2009automated,zhu2012removing,lau2019restoration}. Adaptive optics-based techniques require expensive and complex hardware, hence image processing methods are often preferred. Most image processing techniques are designed for mitigating atmospheric turbulence effects in videos. They \cite{aubailly2009automated,chimitt2019rethinking,anantrasirichai2013atmospheric} often proceeds by combining the complementary clear regions across frames by the lucky fusion process and further performing deconvolution on this fused frame to get the restored output. With the success of deep networks in fast and accurate image reconstruction, a few deep learning techniques have also been proposed for atmospheric turbulence mitigation \cite{yasarla2020learning,yasarla2021learning,lau2021semi,lau2020atfacegan}. All of these methods focus on restoring facial images degraded by atmospheric turbulence. The methods \cite{yasarla2020learning,yasarla2021learning,lau2020atfacegan} have been proposed for single image turbulence mitigation and the method \cite{nair2021confidence} is for turbulence mitigation in videos.

The performance of deep image restoration networks depends on the type of simulation method used for training the network.  In order to gain further insight and also to compare the effectiveness of various simulation methods, in this paper, we study the performance of two state-of-the-art image restoration networks when trained by various turbulence simulation methods.  The performance is evaluated on the real-world LRFID data consisting of face images degraded by turbulence.  We present detailed analysis and comparison of different methods.  


\section{Recent open-source simulation techniques }
\vspace{-0.3cm}
\subsection{Method from Chak \textit{et al.} \cite{chak2021subsampled}}


Inspired by the turbulence formulation in \cite{zhu2012removing,lau2019variational}, Chak \textit{et al.}\cite{chak2021subsampled} use the following mathematical model to simulate the degradation caused by turbulence.
\setlength{\belowdisplayskip}{0pt} \setlength{\belowdisplayshortskip}{0pt}
\setlength{\abovedisplayskip}{0pt} \setlength{\abovedisplayshortskip}{0pt}
\begin{equation}
      T = D(H(I)))+n,
      \label{eq: turb}
 \end{equation}
 where $T$ is the distorted image, $I$ is the clean image, $n$ is additive noise, $D$ is a deformation operator which deforms the image randomly, and $H$ is a  blurring  operator. The deformation operator $D$, is created using the the following process. From the given image $I$, $K$ pixels are selected. For each of these pixels at locations $(x,y)$, a random motion vector $M^{(x,y)}$ of size $S \times S$ is created according to, 
 \begin{equation}
      M^{(x,y)} = \eta(G_{\sigma}*N_1,G_{\sigma}*N_2)
      \label{eq: deform}
 \end{equation}
 where $G_{\sigma}$ is a Gaussian kernel with standard deviation '$\sigma$', '$\eta$' is the strength of the deformation and $N_1$ and $N_2$ are sampled from a standard Gaussian distribution. The deformation operation $D(x)$  and the overall motion vector field over all $K$ points is given by
  \begin{equation}
      M =  \sum_{i=1}^K M^{(x,y)},\;\;  D(x)=M\oplus x,
      \label{eq: deform}
 \end{equation}
where $\oplus$ is the warping operation, the blurring operator $H$ is a Gaussian blur kernel. Table \ref{table:chak} contains the values of hyperparameters used for our experiments.

\begin{table}[tbp!]
\begin{center}
\caption{Parameters of Chak \textit{et al.}\cite{chak2021subsampled}.}
\label{table:chak}
\vspace{-10pt}
\begin{tabular}{ l | c}
\toprule
Patch size (S)& 6 \\ 
Iterations(K) & 1000 + 3000\text{ randint(0,4)}\\ 
Deformation strength ($\eta$) & [0.13 to 0.25]\\
$\sigma$ Range & 16 \\
\bottomrule
\end{tabular}
\end{center}
\end{table}

\vspace{-0.3cm}
\subsection{Method from Schwartzman \textit{et al.} \cite{schwartzman2017turbulence}}

Schwartzman \textit{et al.} \cite{schwartzman2017turbulence} follows a physics-based approach of turbulence modeling. The method introduces an efficient way to render 2D image distortions from physics-based distortion vector fields. The key idea in this method is derived from an empirical model of atmospheric turbulence. Empirically, the distortion motion vectors at two pixels $p$ and $p+v$ are related and their autocorrelation is given by,  
  \begin{equation}
      C(v)=  E[e(p)^Te(p+v)],
      \label{eq: deform}
 \end{equation}
 where $e$ denotes the distortion motion vector function. Now, according to Belen \textit{et al.}\cite{belen2001turbulence}, the function $C(v)$ follows a particular form for turbulence based distortions. This idea is used to traceback to the distortion motion vectors $e(p)$ for all pixels $p$ in the image plane. Once the distortion motion vector function $e$ is obtained, each pixel in the image is warped using this function to create the turbulence distorted image. Table \ref{table:israel} contains the camera and atmospheric parameter settings used in our experiments.
   
\begin{table}[tbp!]
\begin{center}
\caption{Parameters of Schwartzman \textit{et al.}\cite{chak2021subsampled}.}
\vspace{-10pt}
\begin{tabular}{ l | c}
\toprule
Lens Diameter & 0.53 m \\ 
Pixel distance & 4 $\mu m$\\ 
Propagation distance & [2000-5000] m\\
Refractive index parameter$(C_n^2)$  & $3.6e^{-13}$ \\
\bottomrule
\end{tabular}
\label{table:israel}
\end{center}

\vspace{-1cm}
\end{table}
\begin{table*}[t!]
\vspace{-0.5cm}
	\begin{center}
			\caption{Quantitative results for different ranges in  real world turbulence distorted face image dataset(LRFID \cite{miller2019data}). The networks AT-Net\cite{yasarla2021learning} and MPRNet\cite{Zamir2021MPRNet} are trained using synthetic images generated using the five simualtion techniques and tested on LRFID dataset. Green colour highlights the best value over all the simulation methods for  the individual networks.$(\uparrow)$ represents higher the better.
			\label{table:quant}}
			\vspace{-10pt}
		\resizebox{\textwidth}{!}{
			\begin{tabular}{c c c c c c | c c c c c}
				\hline
				&\multicolumn{5}{c}{AT-Net\cite{yasarla2021learning}}&\multicolumn{5}{c}{MPRNet\cite{Zamir2021MPRNet}}\\
				Metric&Chak\cite{chak2021subsampled}&Schwartzman\cite{schwartzman2017turbulence}&Mao\cite{mao2021accelerating}&Chimitt\cite{chimitt2020simulating}&Mei\cite{mei2021ltt}&Chak\cite{chak2021subsampled}&Schwartzman S\cite{schwartzman2017turbulence}&Mao\cite{mao2021accelerating}&Chimitt\cite{chimitt2020simulating}&Mei\cite{mei2021ltt}\\				\hline
				\multicolumn{6}{c}{}&\hspace{-2cm}300m dataset & \multicolumn{3}{c}{}\\
				Top-1$(\uparrow)$&22.47&5.61&14.60&\cellcolor{green!25}24.71&21.35&21.34&17.97&15.37&20.22&\cellcolor{green!25}21.78\\
				Top-3$(\uparrow)$&35.95&13.48&24.71&\cellcolor{green!25}44.94&31.46&31.46&25.84&35.95&\cellcolor{green!25}37.07&36.63\\
				Top-5$(\uparrow)$&42.69&26.96&30.33&\cellcolor{green!25}55.05&46.06&41.57&34.83&42.69&38.20&\cellcolor{green!25}46.53\\
				NIQE$(\downarrow)$&\cellcolor{green!25}5.286&6.584&6.530&6.152&6.768&6.064&6.186&6.150&\cellcolor{green!25}5.248&6.601\\
				BRISQUE$(\downarrow)$&48.85&57.36&61.88&\cellcolor{green!25}37.72&57.91&56.64&61.01&59.46&\cellcolor{green!25}38.72&58.12\\
				\hline
			    \multicolumn{6}{c}{}&\hspace{-2cm}650m dataset & \multicolumn{3}{c}{}\\
                Top-1$(\uparrow)$&0&3.33&3.33&0&6.67&3.33&3.33&3.33&10.00&\cellcolor{green!25}13.33\\  
				Top-3$(\uparrow)$&10.0&13.33&10.00&\cellcolor{green!25}16.67&16.67&16.67&10.0&20.0&13.33&\cellcolor{green!25}20.00\\ 
				Top-5$(\uparrow)$&23.33&13.33&16.67&\cellcolor{green!25}26.67&26.67&30.00&23.33&23.33&\cellcolor{green!25}30.00&20.00\\ 
				NIQE$(\downarrow)$&6.458&7.321&\cellcolor{green!25}5.688&6.327&6.678&6.421&7.627&6.810&\cellcolor{green!25}6.191&6.393\\
				BRISQUE$(\downarrow)$&51.50&65.97&51.28&\cellcolor{green!25}48.52&55.52&58.08&70.27&63.99&\cellcolor{green!25}57.32&59.32\\ 
				\hline
			    \multicolumn{6}{c}{}&\hspace{-2cm}1000m dataset & \multicolumn{3}{c}{}\\
                Top-1$(\uparrow)$&0&0&6.67&6.67&6.67&13.33&13.33&\cellcolor{green!25}20.00&6.667&13.33\\  
				Top-3$(\uparrow)$&26.67&6.67&13.33&\cellcolor{green!25}26.67&20.00&\cellcolor{green!25}33.33&33.33&33.33&33.33&26.66\\ 
				Top-5$(\uparrow)$&40.00&20.00&40.00&\cellcolor{green!25}46.67&46.67&\cellcolor{green!25}46.67&33.33&40.0&40.00&33.33\\  
				NIQE$(\downarrow)$&\cellcolor{green!25}5.647&6.677&6.375&6.529&6.652&6.533&7.434&6.639&\cellcolor{green!25}6.018&6.250\\
				BRISQUE$(\downarrow)$&52.38&60.05&64.46&\cellcolor{green!25}51.62&54.92&57.40&68.57&64.54&\cellcolor{green!25}55.50&58.25\\ \hline
			\end{tabular}
		}
	\end{center}
\end{table*}
\vspace{-0.3cm}
\subsection{Method from Chimitt \textit{et al.} \cite{chimitt2020simulating}} 
Similar to Schwartzman \textit{et al.}\cite{schwartzman2017turbulence}, Chimitt \textit{et al.}\cite{chimitt2020simulating} also follow a physics-based formulation of turbulence. The basis for this work is derived by establishing the equivalence between the angle-of-arrival correlation in Basu \textit{et al.} \cite{basu2015estimation}, and the multi-aperture
correlation by Chanan \cite{chanan1992calculation}. Essentially, Chimitt \textit{et al.} \cite{chimitt2020simulating} first decomposes the geometric distortions (tilts) and blurring effect (aberrations) by utilizing Zernike decomposition. The Zernike coefficients are derived according to the correlation matrix defining the angle of arrival correlations. The tilts are derived from spatial correlation matrix and the blurring effect is drawn from the inter-mode covariance matrix. The overall flow of the simulator is as follows. First, the image is partitioned into blocks and a spatially varying blur estimated using the Zernike coefficients is applied. Then the image is warped according to tilts drawn by utilizing the correlation matrix. These tilts represents the geometric distortion at different pixels in the image. Table \ref{table:chimitt} contains the values of hyperparameters used for our experiments.
\begin{table}[htp!]
\begin{center}
\caption{Parameters of Chimitt \textit{et al.}\cite{chimitt2020simulating}.}
\vspace{-10pt}
\begin{tabular}{ l | c}
\toprule
Aperture Diameter & 0.2034 m \\ 
Wavelength & 525 nm\\ 
Refractive index parameter$(C_n^2)$  & $1e^{-14}$ \\
Focal length & 1.2 m \\
Propogation Length& [300,650,1000] m \\
\bottomrule
\end{tabular}
\label{table:chimitt}
\end{center}
\vspace{-0.7cm}

\end{table}
\vspace{-0.3cm}
\subsection{Method from Mao \textit{et al.} \cite{mao2021accelerating}}
 This paper is inspired from the work from Chimitt \textit{et al.} \cite{chimitt2020simulating}. As can be seen from Chimitt \textit{et al.} \cite{chimitt2020simulating}, estimating the blurring or the aberration operator from the basis of Zernike coefficients is a computationally expensive task compared to the tilt coefficients. To simplify this process, Mao \textit{et al.} \cite{mao2021accelerating} proposed a method to bypass the expensive PSF formation process by learning the mapping between the Zernike coefficients that represents the the phase domain to the spatial domain. For this, a simple neural network is utilized to learn this mapping. Overall the method first reformulates the spatially varying convolution as a set of invariant convolutions and then learns the basis functions for these invariant convolutions using known turbulence statistical models. These convolutions form the aberration operator. The tilt coefficients are computed in an inexpensive manner. Finally, both the tilts and the abberations are used to model turbulence distortion in the image. Table \ref{table:mao} contains the hyperparameters used in our experiments.
\vspace{-10pt}
\begin{table}[htp!]
\begin{center}
\caption{Parameters of Mao \textit{et al.}\cite{chimitt2020simulating}.}
\vspace{-10pt}
\begin{tabular}{ l | c}
\toprule
Aperture Diameter & 0.1 m\\ 
Fried parameter & 0.02\\ 
Wavelength  & 500 nm \\
Propagation Length& [300,650,1000] m \\
Distortion strength & 5 \\
\bottomrule
\end{tabular}
\label{table:mao}
\end{center}
\end{table}


 \begin{figure}[t!]
    \centering
    \includegraphics[width=\linewidth]{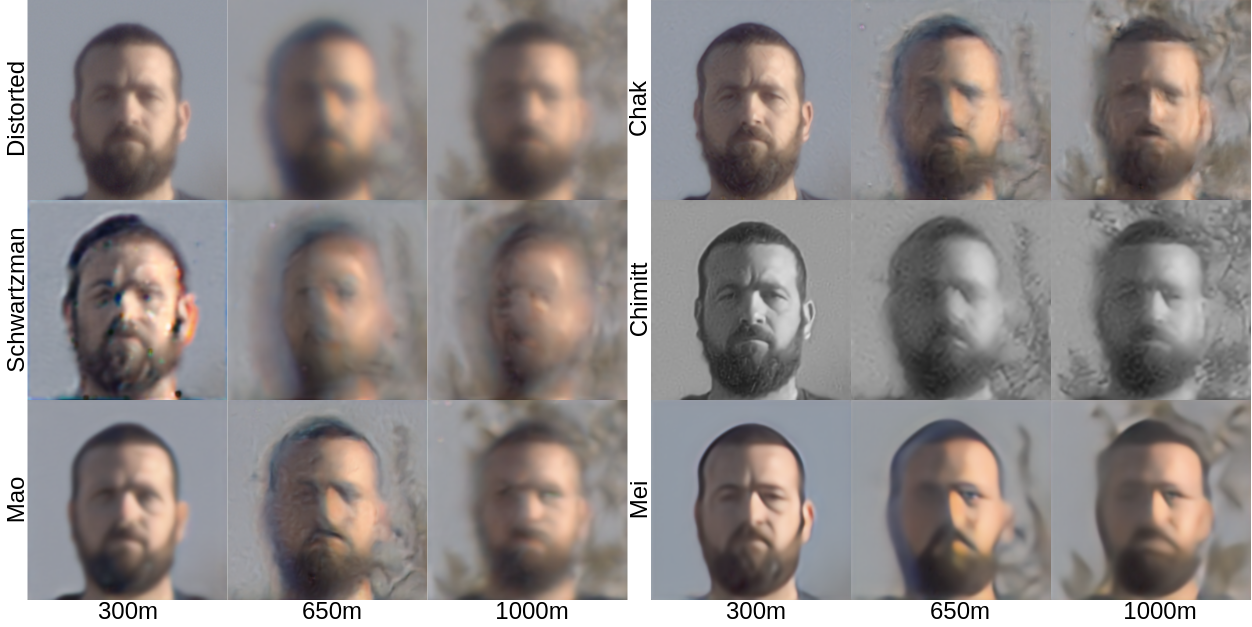}
    \includegraphics[width=\linewidth]{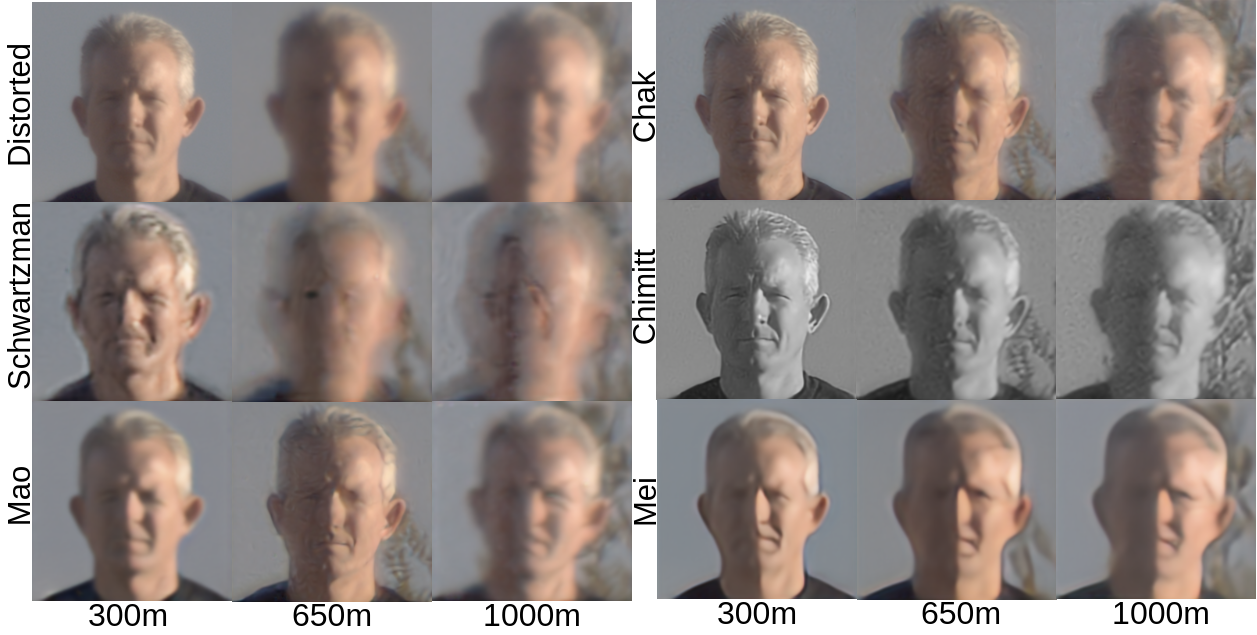}
   \vspace{-15pt} \caption{Restored Images from LRFID dataset using AT-Net\cite{yasarla2021learning}. The methods used for preparinng the training datasets are:  First row:  Distorted and Chak et al \cite{chak2021subsampled}. Second row: Schwartzman \textit{et al.} \cite{schwartzman2017turbulence} and Chimitt \textit{et al.} \cite{chimitt2020simulating} and .Third row: Mao \textit{et al.}\cite{mao2021accelerating} and \textit{Mei et al.}\cite{mei2021ltt}.}
    \label{fig:quali}
\end{figure}
\vspace{-.5cm}
\subsection{Method from Mei and Patel \cite{mei2021ltt}}

Recently, Mei and Patel \cite{mei2021ltt} proposed a method that can model higher-order distortions due to turbulence using a simple method, called ElasticAug. The mathematical model for this method is the same as that in Equation \ref{eq: turb}. This method uses  Gaussian blur as the blurring operator and uses the elastic transformation \cite{simard2003best} as the deformation operator. The elastic transformation displaces pixels using random motion vectors. Table \ref{table:mei} contains the parameters used for the blur and elastic transformations in our experiments

\begin{table}[tb!]
\begin{center}
\caption{Parameters of ElasticAug.}
\vskip-10pt
\begin{tabular}{ l | c}
\toprule
Blur Kernel Size & 41 \\ 
Blur Kernel Types & [Isotropic, Anisotropic ]\\ 
Blur $\sigma$ Range & [1 to 25] \\
Down-sample Range & [$\frac{1}{8}$ to 1] \\
Elastic $\alpha$ Range & [0 to 50] \\
Elastic $\sigma$ Range & [4 to 5] \\
\bottomrule
\end{tabular}
\label{table:mei}
\end{center}
\vspace{-0.7cm}
\end{table}

\section{Image RESTORATION NETWORKS}

We evaluate the effectiveness of the simulation methods using two state-of-the-art deep networks \cite{yasarla2021learning,Zamir2021MPRNet}. 


\noindent {\bf{Turbulence Distortion Removal Network (AT-Net) \cite{yasarla2021learning}.}} The first network that we use is the state-of-the-art network for single image turbulence mitigation. AT-Net \cite{yasarla2021learning} first captures the regions that are hard to restore using epistemic uncertainty which are estimated based on Monte Carlo simulations. A restoration network is then guided using the estimated uncertainty maps to obtain the restored image. For all the simulation techniques, we re-train AT-Net with the following parameters. The number of Monte-Carlo simulations $S=10$, Learning rate $=2\times 10^{-4}$, batch size $b=2$, number of epochs $=32$ and we use Adam optimizer\cite{kingma2014adam}.


\noindent {\bf{Multi-Stage Progressive Image Restoration (MPRNet) \cite{Zamir2021MPRNet}.}} The second network we use is the state-of-the-art network for general image restoration. MPRNet \cite{Zamir2021MPRNet} progressively learns restoration functions for degraded images through a multi-scale architecture. For the lower scales, contextualized features are learned using encoder-decoder architectures, which are later fused with higher resolutions. At each scale a pixel-wise attention mechanism is used to re-weight the features. The parameters used for MPRNet are: batchsize$=2$, learning rate = $2 \times 10^{-4}$.  We use Adam optimizer\cite{kingma2014adam} for the training process. 
\section{Experimental Results}

In this section, we present the performance of AT-Net \cite{yasarla2021learning} and MPRNet \cite{Zamir2021MPRNet} trained using different simulation methods. We train both networks by augmenting distortions produced by  different simulation methods on 10,000 randomly selected images from the FFHQ dataset \cite{karras2019style}. We use images of size $256\times 256$ obtained by re-scaling clean images from the FFHQ dataset for training. Both networks are trained using their default loss functions, as mentioned in their respective papers. We first re-train both networks using synthetic images generated using the simulation methods and test the performance of these trained networks on the LRFID dataset \cite{miller2019data}.

\noindent {\bf{Evaluation dataset.}}  The LRFID dataset \cite{miller2019data} consists of real-world turbulence distorted images of 100 individuals in 6 different poses. These images are captured at different ranges -- 300 meters, 650 meters, and 1000 meters. The atmospheric properties of the background conditions in the captured images can be seen in Table \ref{table:atmospheric}. Sample turbulence distorted images can be found in the third row of Fig.\ref{fig:combined}. The LRFID dataset doesn't have paired ground truth images in the same setting; rather, it contains a gallery set of the same 100 individuals in indoor conditions. As we can see from Fig. \ref{fig:combined}, the distortion level is quite high for the 1000m range. This holds true for most of the images at ranges 650 meters also.  Hence only a very few images are present in the testing dataset prepared for evaluation.
 
 \begin{table}[tp!]
 	\caption{Details of the atmospheric properties of the videos in the real-world dataset.}
 	 	\vspace{-0.5cm}
 	\small
 	\begin{center}
 		\resizebox{0.5\textwidth}{!}{
 			\begin{tabular}{|p{15mm}|p{18mm}|p{15mm}|p{20mm}|p{15mm}|p{20mm}|p{18mm}|}
 				\hline
 				Atmospheric \newline Properties & Distance \newline from camera \newline (meters) & temperature \newline ( $^{\circ}$ Celsius) & wind speed \newline (meters/second) &Humidity \newline  ($\%$) & Wind direction \newline (radians) & Atmospheric \newline pressure \newline (millibar)\\
 				\hline\
 				Values&300-600& 16.5-38.8 &0.57-9.34& 5-28.5  &0-2$\pi$ & 919.3- 931.2 \\
 				\hline
 			\end{tabular}
 		}
 	\end{center}
 	\label{table:atmospheric}
 	\vspace{-0.5cm}
 \end{table}


\noindent {\bf{Metrics.}}   Since the corresponding clean target images in the LRFID dataset \cite{miller2019data} are not available, the performance of the simulation methods are evaluated using the facial recognition scores obtained on the restored images. Specifically, we use the Top-1, Top-3, and Top-5 facial recognition scores. The Top-K score refers to the actual identity being present in the K nearest matches from the gallery set. For obtaining the recognition scores, we use  Arcface \cite{deng2019arcface} facial recognition algorithm by using the clean images of the people in the LRFID dataset \cite{miller2019data} in the indoor conditions as the gallery set. For fair evaluations, all images are converted to grayscale before estimating the recognition scores. We also evaluate the methods using no-reference image quality metrics such as natural image quality evaluator (NIQE) \cite{mittal2012making} scores and blind/reference less
image spatial quality evaluator (BRISQUE) scores \cite{mittal2011blind}. NIQE and BRISQUE scores represent the naturalness of the images. Smaller scores represent better perceptual quality of the images.   



\noindent {\bf{Results.}} The quantitative results are presented in Table \ref{table:quant}  and the corresponding qualitative results can be found in Fig. \ref{fig:quali}. In the case of 300m, as can be seen from Table \ref{table:atmospheric}, all turbulence simulation methods perform well. The best facial recognition scores for AT-Net  are obtained by using Chimitt  \textit{et al.}  \cite{chimitt2020simulating} as the simulation method for generating synthetic training data. For MPRNet, Chimitt \textit{et al.} \cite{chimitt2020simulating} gives the best results for Top-3 recognition score, NIQE, and BRISQUE scores, while the best Top-1 and Top-5 recognition scores are obtained by using the simulation technique from Mei and Patel \cite{mei2021ltt}. From the  qualitative results in Fig.~\ref{fig:quali}, we can see that the networks when trained using Chimitt \textit{et al.} \cite{chimitt2020simulating}, produce the most visually pleasing results for the images at 300m. Although the method from Mao \textit{et al.} \cite{mao2021accelerating} is a more recent technique for simulating atmospheric turbulence, the blurring aberrations in Mao \textit{et al.}\cite{mao2021accelerating} are computed using the blurring aberrations derived in the method by Chimitt \textit{et al.} \cite{chimitt2020simulating}. Hence being an approximation of Chimitt \textit{et al.} \cite{chimitt2020simulating}. This is the main reason why the performance of the simulation method from Mao \textit{et al.} \cite{mao2021accelerating} is lower for almost all evaluations when compared to the method from Chimitt  \textit{et al.} \cite{chimitt2020simulating}. For the method from Schwartzman \textit{et al.} \cite{schwartzman2017turbulence}, as we can see from Fig. \ref{fig:combined}, the tilt model is good, but the aberration model doesn't match with that of the real-world turbulence images resulting in low recognition scores as can be seen in Fig.~ \ref{fig:quali}. Hence an effective simulation method for preparing training data to train deep networks for turbulence mitigation in grayscale images would be the method from Chimitt \textit{et al.} \cite{chimitt2020simulating}, and for color images would be the method from Mei and Patel \cite{mei2021ltt}. 

As can be seen from Eq.~\ref{eq: turb}, turbulence mitigation is an ill-posed problem. Thus, restoring the exact facial identity is difficult when much of the salient facial features are not present in the distorted image. Hence, restoring images at higher ranges is very difficult while accurately reconstructing the facial features. This is the reason why the Top-1 recognition scores for the simulation methods are low for the higher ranges, which can be seen in Table \ref{table:quant}. 
For AT-Net\cite{yasarla2021learning}, for both 650m  and 1000m, the method from Chimitt \textit{et al.} \cite{chimitt2020simulating} gives the best facial recognition and BRISQUE scores.  One drawback of the method from Chimitt \textit{et al.} \cite{chimitt2020simulating} is that it is only defined for grayscale images. For color images, we have found that the method from Mei and Patel \cite{mei2021ltt} utilizing elastic augmentations works the best for 650m range. While the method from Chak \textit{et al.} \cite{chak2021subsampled} performs the best for 1000m range. One key observation from Table~\ref{table:quant} is that the quantitative evaluation metrics from all simulation methods are very close for ranges 650m and 1000m. This is because the test set size is quite small and also the amount of distortion present in these images is quite high.  Hence for the higher ranges, judging from the quantitative values in Table\ref{table:quant}, the turbulence phenomenon is best modelled by Chimitt \textit{et al.} for grayscale images. In the case of colour images, since the values are very close, any one of the simulation methods from Chak\textit{et al.}\cite{chak2021subsampled}, Mao \textit{et al.}\cite{mao2021accelerating} or Mei and Patel\cite{mei2021ltt} could be used.



\section{CONCLUSION}
In this paper, we have studied the effectiveness of five recent atmospheric turbulence simulation methods on image restoration.  Two sate-of-the-art image restoration networks were utilized in our study.  Based on facial recognition scores and no-reference image quality methods, we observe that the method from Chimitt \textit{et al.} \cite{chimitt2020simulating}  gives the best possible simulated images in the case of grayscale images, and the technique from Mei and Patel \cite{mei2021ltt} gives the best possible images in the case of RGB images.
\clearpage
\bibliographystyle{IEEEbib}
{\footnotesize
\bibliography{strings,refs}

\begin{thebibliography}{10}

\bibitem{lau2019restoration}
Chun~Pong Lau, Yu~Hin Lai, and Lok~Ming Lui,
\newblock ``Restoration of atmospheric turbulence-distorted images via rpca and
  quasiconformal maps,''
\newblock {\em Inverse Problems}, vol. 35, no. 7, pp. 074002, 2019.

\bibitem{mei2021ltt}
Kangfu Mei and Vishal~M Patel,
\newblock ``Ltt-gan: Looking through turbulence by inverting gans,''
\newblock {\em arXiv preprint arXiv:2112.02379}, 2021.

\bibitem{yasarla2021learning}
Rajeev Yasarla and Vishal~M Patel,
\newblock ``Learning to restore images degraded by atmospheric turbulence using
  uncertainty,''
\newblock in {\em 2021 IEEE International Conference on Image Processing
  (ICIP)}. IEEE, 2021, pp. 1694--1698.

\bibitem{deng2019arcface}
Jiankang Deng, Jia Guo, Niannan Xue, and Stefanos Zafeiriou,
\newblock ``Arcface: Additive angular margin loss for deep face recognition,''
\newblock in {\em Proceedings of the IEEE/CVF conference on computer vision and
  pattern recognition}, 2019, pp. 4690--4699.

\bibitem{deng2020retinaface}
Jiankang Deng, Jia Guo, Evangelos Ververas, Irene Kotsia, and Stefanos
  Zafeiriou,
\newblock ``Retinaface: Single-shot multi-level face localisation in the
  wild,''
\newblock in {\em Proceedings of the IEEE/CVF Conference on Computer Vision and
  Pattern Recognition}, 2020, pp. 5203--5212.

\bibitem{nair2021confidence}
Nithin~Gopalakrishnan Nair and Vishal~M Patel,
\newblock ``Confidence guided network for atmospheric turbulence mitigation,''
\newblock in {\em 2021 IEEE International Conference on Image Processing
  (ICIP)}. IEEE, 2021, pp. 1359--1363.

\bibitem{yasarla2020learning}
Rajeev Yasarla and Vishal~M Patel,
\newblock ``Learning to restore a single face image degraded by atmospheric
  turbulence using cnns,''
\newblock {\em arXiv preprint arXiv:2007.08404}, 2020.

\bibitem{chak2021subsampled}
Wai~Ho Chak, Chun~Pong Lau, and Lok~Ming Lui,
\newblock ``Subsampled turbulence removal network,''
\newblock {\em Mathematics, Computation and Geometry of Data}, vol. 1, no. 1,
  pp. 1--33, 2021.

\bibitem{hardie2017simulation}
Russell~C Hardie, Jonathan~D Power, Daniel~A LeMaster, Douglas~R Droege, Szymon
  Gladysz, and Santasri Bose-Pillai,
\newblock ``Simulation of anisoplanatic imaging through optical turbulence
  using numerical wave propagation with new validation analysis,''
\newblock {\em Optical Engineering}, vol. 56, no. 7, pp. 071502, 2017.

\bibitem{potvin2011simple}
Guy Potvin, J~Luc Forand, and Denis Dion,
\newblock ``A simple physical model for simulating turbulent imaging,''
\newblock in {\em Infrared Imaging Systems: Design, Analysis, Modeling, and
  Testing XXII}. International Society for Optics and Photonics, 2011, vol.
  8014, p. 80140Y.

\bibitem{mao2021accelerating}
Zhiyuan Mao, Nicholas Chimitt, and Stanley~H Chan,
\newblock ``Accelerating atmospheric turbulence simulation via learned
  phase-to-space transform,''
\newblock in {\em Proceedings of the IEEE/CVF International Conference on
  Computer Vision}, 2021, pp. 14759--14768.

\bibitem{chimitt2020simulating}
Nicholas Chimitt and Stanley~H Chan,
\newblock ``Simulating anisoplanatic turbulence by sampling intermodal and
  spatially correlated zernike coefficients,''
\newblock {\em Optical Engineering}, vol. 59, no. 8, pp. 083101, 2020.

\bibitem{schmidt2010numerical}
Jason~Daniel Schmidt,
\newblock ``Numerical simulation of optical wave propagation: With examples in
  matlab,''
\newblock SPIE, 2010.

\bibitem{bos2012technique}
Jeremy~P Bos and Michael~C Roggemann,
\newblock ``Technique for simulating anisoplanatic image formation over long
  horizontal paths,''
\newblock {\em Optical Engineering}, vol. 51, no. 10, pp. 101704, 2012.

\bibitem{schwartzman2017turbulence}
Armin Schwartzman, Marina Alterman, Rotem Zamir, and Yoav~Y Schechner,
\newblock ``Turbulence-induced 2d correlated image distortion,''
\newblock in {\em 2017 IEEE International Conference on Computational
  Photography (ICCP)}. IEEE, 2017, pp. 1--13.

\bibitem{roggemann1996imaging}
Michael~C Roggemann, Byron~M Welsh, and Bobby~R Hunt,
\newblock {\em Imaging through turbulence},
\newblock CRC press, 1996.

\bibitem{pearson1976atmospheric}
James~E Pearson,
\newblock ``Atmospheric turbulence compensation using coherent optical adaptive
  techniques,''
\newblock {\em Applied optics}, vol. 15, no. 3, pp. 622--631, 1976.

\bibitem{tyson2015principles}
Robert~K Tyson,
\newblock {\em Principles of adaptive optics},
\newblock CRC press, 2015.

\bibitem{anantrasirichai2013atmospheric}
Nantheera Anantrasirichai, Alin Achim, Nick~G Kingsbury, and David~R Bull,
\newblock ``Atmospheric turbulence mitigation using complex wavelet-based
  fusion,''
\newblock {\em IEEE Transactions on Image Processing}, vol. 22, no. 6, pp.
  2398--2408, 2013.

\bibitem{anantrasirichai2018atmospheric}
Nantheera Anantrasirichai, Alin Achim, and David Bull,
\newblock ``Atmospheric turbulence mitigation for sequences with moving objects
  using recursive image fusion,''
\newblock in {\em 2018 25th IEEE International Conference on Image Processing
  (ICIP)}. IEEE, 2018, pp. 2895--2899.

\bibitem{aubailly2009automated}
Mathieu Aubailly, Mikhail~A Vorontsov, Gary~W Carhart, and Michael~T Valley,
\newblock ``Automated video enhancement from a stream of
  atmospherically-distorted images: the lucky-region fusion approach,''
\newblock in {\em Atmospheric Optics: Models, Measurements, and
  Target-in-the-Loop Propagation III}. International Society for Optics and
  Photonics, 2009, vol. 7463, p. 74630C.

\bibitem{zhu2012removing}
Xiang Zhu and Peyman Milanfar,
\newblock ``Removing atmospheric turbulence via space-invariant
  deconvolution,''
\newblock {\em IEEE transactions on pattern analysis and machine intelligence},
  vol. 35, no. 1, pp. 157--170, 2012.

\bibitem{chimitt2019rethinking}
Nicholas Chimitt, Zhiyuan Mao, Guanzhe Hong, and Stanley~H Chan,
\newblock ``Rethinking atmospheric turbulence mitigation,''
\newblock {\em arXiv preprint arXiv:1905.07498}, 2019.

\bibitem{lau2021semi}
Chun~Pong Lau, Amit Kumar, and Rama Chellappa,
\newblock ``Semi-supervised landmark-guided restoration of atmospheric
  turbulent images,''
\newblock {\em IEEE Journal of Selected Topics in Signal Processing}, vol. 15,
  no. 2, pp. 204--215, 2021.

\bibitem{lau2020atfacegan}
Chun~Pong Lau, Hossein Souri, and Rama Chellappa,
\newblock ``Atfacegan: Single face image restoration and recognition from
  atmospheric turbulence,''
\newblock in {\em 2020 15th IEEE International Conference on Automatic Face and
  Gesture Recognition (FG 2020)}. IEEE, 2020, pp. 32--39.

\bibitem{lau2019variational}
Chun~Pong Lau, Yu~Hin Lai, and Lok~Ming Lui,
\newblock ``Variational models for joint subsampling and reconstruction of
  turbulence-degraded images,''
\newblock {\em Journal of Scientific Computing}, vol. 78, no. 3, pp.
  1488--1525, 2019.

\bibitem{belen2001turbulence}
Mikhail~S Belen’kii, John~M Stewart, and Patti Gillespie,
\newblock ``Turbulence-induced edge image waviness: theory and experiment,''
\newblock {\em Applied optics}, vol. 40, no. 9, pp. 1321--1328, 2001.

\bibitem{miller2019data}
Kevin~J Miller, Bradley Preece, Todd~W Du~Bosq, and Kevin~R Leonard,
\newblock ``A data-constrained algorithm for the emulation of long-range
  turbulence-degraded video,''
\newblock in {\em Infrared Imaging Systems: Design, Analysis, Modeling, and
  Testing XXX}. International Society for Optics and Photonics, 2019, vol.
  11001, p. 110010J.

\bibitem{Zamir2021MPRNet}
Syed~Waqas Zamir, Aditya Arora, Salman Khan, Munawar Hayat, Fahad~Shahbaz Khan,
  Ming-Hsuan Yang, and Ling Shao,
\newblock ``Multi-stage progressive image restoration,''
\newblock in {\em CVPR}, 2021.

\bibitem{basu2015estimation}
Santasri Basu, Jack~E McCrae, and Steven~T Fiorino,
\newblock ``Estimation of the path-averaged atmospheric refractive index
  structure constant from time-lapse imagery,''
\newblock in {\em Laser Radar Technology and Applications XX; and Atmospheric
  Propagation XII}. International Society for Optics and Photonics, 2015, vol.
  9465, p. 94650T.

\bibitem{chanan1992calculation}
Gary~A Chanan,
\newblock ``Calculation of wave-front tilt correlations associated with
  atmospheric turbulence,''
\newblock {\em JOSA A}, vol. 9, no. 2, pp. 298--301, 1992.

\bibitem{simard2003best}
Patrice~Y Simard, David Steinkraus, John~C Platt, et~al.,
\newblock ``Best practices for convolutional neural networks applied to visual
  document analysis.,''
\newblock in {\em Icdar}, 2003, vol.~3.

\bibitem{kingma2014adam}
Diederik~P Kingma and Jimmy Ba,
\newblock ``Adam: A method for stochastic optimization,''
\newblock {\em arXiv preprint arXiv:1412.6980}, 2014.

\bibitem{karras2019style}
Tero Karras, Samuli Laine, and Timo Aila,
\newblock ``A style-based generator architecture for generative adversarial
  networks,''
\newblock in {\em Proceedings of the IEEE/CVF Conference on Computer Vision and
  Pattern Recognition}, 2019, pp. 4401--4410.

\bibitem{mittal2012making}
Anish Mittal, Rajiv Soundararajan, and Alan~C Bovik,
\newblock ``Making a “completely blind” image quality analyzer,''
\newblock {\em IEEE Signal processing letters}, vol. 20, no. 3, pp. 209--212,
  2012.

\bibitem{mittal2011blind}
Anish Mittal, Anush~K Moorthy, and Alan~C Bovik,
\newblock ``Blind/referenceless image spatial quality evaluator,''
\newblock in {\em 2011 conference record of the forty fifth asilomar conference
  on signals, systems and computers (ASILOMAR)}. IEEE, 2011, pp. 723--727.

\end{thebibliography}
}
\end{document}